\documentclass{ecai} 
\usepackage{latexsym}
\usepackage{amssymb}
\usepackage{amsmath}
\usepackage{amsthm}
\usepackage{booktabs}
\usepackage{enumitem}
\usepackage{graphicx}
\usepackage{color}
\usepackage{algorithm}
\usepackage{algorithmic}
\usepackage{stfloats}
\usepackage{siunitx}
\usepackage{epsfig}
\usepackage{multirow}
\usepackage[table]{xcolor}
\usepackage{float}
\definecolor{grayrow}{cmyk}{0,0,0,0.18}
\usepackage{hyperref}
\usepackage[justification=justified]{caption}
\newcommand{\BibTeX}{B\kern-.05em{\sc i\kern-.025em b}\kern-.08em\TeX}

\begin{document}

%%%%%%%%%%%%%%%%%%%%%%%%%%%%%%%%%%%%%%%%%%%%%%%%%%%%%%%%%%%%%%%%%%%%%%%%

\begin{frontmatter}

%%% Use this command to specify your submission number.
%%% In doubleblind mode, it will be printed on the first page.

\paperid{211} 

%%% Use this command to specify the title of your paper.

\title{Quater-GCN: Enhancing 3D Human Pose Estimation with Orientation and Semi-supervised Training}

%%% Use this combinations of commands to specify all authors of your 
%%% paper. Use \fnms{} and \snm{} to indicate everyone's first names 
%%% and surname. This will help the publisher with indexing the 
%%% proceedings. Please use a reasonable approximation in case your 
%%% name does not neatly split into "first names" and "surname".
%%% Specifying your ORCID digital identifier is optional. 
%%% Use the \thanks{} command to indicate one or more corresponding 
%%% authors and their email address(es). If so desired, you can specify
%%% author contributions using the \footnote{} command.

% \author[A]{\fnms{First}~\snm{Author}\orcid{....-....-....-....}\thanks{Corresponding Author. Email: somename@university.edu.}\footnote{Equal contribution.}}
% \author[B]{\fnms{Second}~\snm{Author}\orcid{....-....-....-....}\footnotemark}
% \author[B,C]{\fnms{Third}~\snm{Author}\orcid{....-....-....-....}} 

% \address[A]{Short Affiliation of First Author}
% \address[B]{Short Affiliation of Second Author and Third Author}
% \address[C]{Short Alternate Affiliation of Third Author}

\author{\fnms{Xingyu}~\snm{Song}}
\author{\fnms{Zhan}~\snm{Li}}
\author{\fnms{Shi}~\snm{Chen}} 
\author{\fnms{Kazuyuki}~\snm{Demachi}} 

\address{The University of Tokyo}
\address{songxingyu0429@gmail.com, \{lizhan, shichen, yypr9411\}@g.ecc.u-tokyo.ac.jp}

%%% Use this environment to include an abstract of your paper.

\begin{abstract}
3D human pose estimation is a vital task in computer vision, involving the prediction of human joint positions from images or videos to reconstruct a skeleton of a human in three-dimensional space.
This technology is pivotal in various fields, including animation, security, human-computer interaction, and automotive safety, where it promotes both technological progress and enhanced human well-being.
The advent of deep learning significantly advances the performance of 3D pose estimation by incorporating temporal information for predicting the spatial positions of human joints. 
However, traditional methods often fall short as they primarily focus on the spatial coordinates of joints and overlook the orientation and rotation of the connecting bones, which are crucial for a comprehensive understanding of human pose in 3D space.
To address these limitations, we introduce Quater-GCN (Q-GCN), a directed graph convolutional network tailored to enhance pose estimation by orientation. 
Q-GCN excels by not only capturing the spatial dependencies among node joints through their coordinates but also integrating the dynamic context of bone rotations in 2D space. 
This approach enables a more sophisticated representation of human poses by also regressing the orientation of each bone in 3D space, moving beyond mere coordinate prediction.
Furthermore, we complement our model with a semi-supervised training strategy that leverages unlabeled data, addressing the challenge of limited orientation ground truth data.
Through comprehensive evaluations, Q-GCN has demonstrated outstanding performance against current state-of-the-art methods.
% The full version of this paper, along with the code and data, can be found at~\cite{song2024quatergcnenhancing3dhuman}.
The code and models of Q-GCN are available at \href{https://github.com/xingyu-song/q_gcn}{here}.
\end{abstract}

\end{frontmatter}

%%%%%%%%%%%%%%%%%%%%%%%%%%%%%%%%%%%%%%%%%%%%%%%%%%%%%%%%%%%%%%%%%%%%%%%%

\section{Introduction}
3D human pose estimation is a critical task aimed at predicting the spatial positions of human joints within three-dimensional space.
This process is foundational for various applications, including action recognition~\cite{model12_stgcn, ar_ddgcn}, synthetic data augmentation~\cite{surreal}, and 3D reconstruction~\cite{SMPL-X, smpl}.
The task of 3D human pose estimation can be categorized based on viewpoint settings into multi-view and monocular views. 
The monocular view, in comparison to the multi-view approach, offers advantages in terms of lower equipment costs and greater flexibility for real-world applications. 
However, the reduced accuracy of monocular estimation presents significant challenges, thereby garnering increased interest.

Recently, the development of deep learning model has significantly improved the outcomes of 3D human pose estimation~\cite{pe_glagcn, jointformer, 3D_PE_semi, RN011}. 
Current deep learning-based techniques fall into two main categories: one-stage methods~\cite{one_stage_Context_modeling, one_stage_I2l_meshnet, one_stage_Probabilistic_monocular} and 2D-to-3D methods~\cite{RN013, RN011, 3D_PE_semi}. 
One-stage methods directly derive the 3D coordinates to construct the human pose from images in an end-to-end manner. 
Conversely, 2D-to-3D methods initiate by estimating the 2D pose from the input image, which is then extrapolated into 3D space. 
Comparing to one-stage methods, the superior performance of 2D-to-3D methods is attributed to advancements in 2D human pose detection and the leveraging of temporal information across multiple frames. 
Moreover, the intermediate 2D pose estimation stage significantly lowers the data volume and simplifies the complexity of the 3D estimation task. 
Thus, we are motivated to explore the potential of enhancing 3D human pose prediction through high-quality estimated 2D pose data.

However, traditional deep learning models for 2D-to-3D pose lifting focus on the spatial coordinates of joints solely which do not explicitly model the orientation or the rotation of the bones connecting these joints. 
This orientation information is crucial for understanding the pose in three-dimensional space, as it provides insights into the direction in which a limb is facing or moving.
In addition, for complex scenarios where multiple body parts are closely interacting or occluded, the lack of orientation information can lead to ambiguities that are difficult to resolve based on position information alone. 
These ambiguities can result in less robust pose estimations in challenging situations (see Section~\ref{sec:results} for details).

Therefore, determining the optimal model architecture to encode the structural information of human body, including both position and orientation information, is our primary focus.
Within the domain of 2D-to-3D pose estimation approaches, recent advancements in deep learning can be categorized into three distinct architectures: Temporal Convolutional Networks (TCN)-based architectures~\cite{3D_PE_semi, RN013}, Transformer architectures~\cite{trans_Mhformer, trans_Seq2seq, trans_temporal_contexts}, and Graph Convolutional Networks (GCN)-based architectures~\cite{pe_glagcn, CDGCN, pr_semgcn}. 
Comparing to another two kinds of architectures, GCN-based architecture stand out for their ability to explicitly maintain the structural integrity of both 2D and 3D human poses throughout the convolutional process, and a more parameter-efficient process, especially when dealing with graph-structured data both on coordinate and orientation. 
These distinctive capabilities of GCN-based models to conserve the pose structure during estimation highlights their potential for achieving more refined outcomes in 3D pose estimation tasks (see Section~\ref{sec:relatedwork} for details).

In this paper, we introduce an innovative Directed Graph Convolutional Network, named Quater-GCN (Q-GCN).
The Q-GCN not only captures spatial dependencies by analyzing the positions of each node joint but also integrates dynamic context through examining the rotation of each bone in the 2D space. 
Similarly, as for pose construction, our method extends beyond solely predicting the coordinates of each joint, it also infers the orientation of each bone in 3D space, crafting a more sophisticated pose representation.

However, regressing the prediction using insufficient orientation ground truth proves challenging.
Additionally, calculating the 4D Orientation of each bone joint using only the 3D coordinates of each joint node can be challenging and typically yields incomplete results due to the absence of directional information in space~\cite{6D}. 
Moreover, gathering precise orientation data typically requires an expensive motion capture setup~\cite{3D_PE_semi}.
Therefore, we have developed a semi-supervised training strategy that effectively uses unlabeled data by mapping the predicted 4D Orientations back to the rotations in 2D space of each bone joint.

The primary contributions of our work can be summarized in three key aspects:
(1) The introduction of a distinctive 2D-to-3D pose lifting method that incorporates bone joint orientations, significantly enhancing model performance.
(2) The development of a semi-supervised training approach, ingeniously leveraging unlabeled data to overcome the scarcity of orientation training data.
(3) Demonstrated improvements in 3D pose estimation accuracy over existing state-of-the-art methods.
% affirming the effectiveness of orientation and semi-supervision strategy in representing a more sophisticated 3D pose.

%%%%%%%%%%%%%%%%%%%%%%%%%%%%%%%%%%%%%%%%%%%%%%%%%%%%%%%%%%%%%%%%%%%%%%%%
\section{Related Work}
\label{sec:relatedwork}
\subsection{2D-to-3D Pose Lifting}
Recent advancements in 2D-to-3D pose lifting models can be categorized into three primary types: TCN-, Transformer-, and GCN-based architectures. 
TCN and Transformer methods are known for their broad receptive fields, effectively processing long 2D pose sequences through strided convolutions. 
TCN-based methods like~\cite{3D_PE_semi, RN013} have enhanced pose lifting with their architecture designs. 
For example,~\cite{3D_PE_semi} introduces a semi-supervised technique called back-projection that uses unlabeled video data to boost accuracy.
\cite{RN013} incorporates an attention mechanism and multi-scale dilated convolutions to address temporal inconsistency and improve accuracy by focusing on key frames.
Transformer-based approaches~\cite{trans_Mhformer, trans_Seq2seq, trans_temporal_contexts} also show promise by utilizing strided structures to handle depth ambiguities and improve spatial and temporal feature encoding. 
This technique enables these models to cover the full video sequence, enhancing the potential for accurate pose estimation.

In contrast, GCN-based models are superior in maintaining the structural integrity of both 2D and 3D poses, preserving joint relationships and offering parameter efficiency, especially with graph-structured data. 
The foundational work on GCNs~\cite{gcn} and further developments like ST-GCN~\cite{model12_stgcn}, which introduced spatial temporal graph convolution for action recognition, and DGCN~\cite{DGCN}, which models skeleton data as a directed acyclic graph, highlight their capabilities. 
Recent implementations in pose lifting~\cite{pe_glagcn, CDGCN, pr_semgcn, ugcn} demonstrate GCNs' versatility. 
Notably,~\cite{CDGCN} and~\cite{pe_glagcn} explore advanced graph convolutional techniques that address dynamic joint dependencies and enhance local feature detection, while~\cite{pr_semgcn} focuses on learning semantic relationships between nodes.

\subsection{Orientation-based Motion Representation}
The concept of orientation first emerged in human motion representation through mesh generation tasks. 
The Skinned Multi-Person Linear model (SMPL)~\cite{smpl} serves as a foundational approach in human motion capture, introducing orientation-based representation of the human body. 
Inspired by SMPL, subsequent models such as CAPE~\cite{shape_cape}, MANO~\cite{shape_MANO:SIGGRAPHASIA:2017}, SMPL-X~\cite{SMPL-X}, and STAR~\cite{shape_STAR:2020} have extended the framework to include detailed body shape modeling, facial expressions, hand movements, and the depiction of clothed human figures.

In the realm of non-parametric models, OriNet~\cite{OriNet} employs limb orientations to depict 3D poses, coupling the orientation with the bounding box of each limb region to enhance the correlation between images and predictions. 
Yet, the orientation error does not regress through iteration.
\cite{6D} presents an innovative method for estimating the complete position and rotation of skeletal joints. 
It utilizes virtual markers to provide ample data, allowing for the accurate deduction of rotations with straightforward post-processing steps. 

In 3D human pose estimation task,~\cite{occlusion_3dHPE_orientation} first proposes a framework based on Mask Region-based Convolutional Neural Networks (R-CNN) and extended to integrate the joint feature, body boundary, body orientation and occlusion condition together.
PONet~\cite{PONet_3dHPE_orientation} then estimates the 4D Orientation of these limbs by taking advantage of the local image evidence to recover the 3D pose.
Similarly, PedRecNet~\cite{PedPecNet_3dHPE_orientation} supports body and head orientation estimation based on full body bounding box input. 
% PoseGraphNet++\cite{PoseGraphNet} proposes an approach to estimate the 3D human skeletal representation by leveraging the mutual geometrical relationships between joint positions and bone orientations, using a GCN to regress both elements. 
% However, their method has two main drawbacks: (1) Both position and orientation are represented relative to either the root joint or the parent bone, while the human skeleton is modeled as an undirected graph. 
% This approach does not adequately capture the articulated nature of human skeletons, as it fails to clearly present the hierarchical relationships among the joints\cite{CDGCN}. 
% (2) The use of a 6D matrix to represent orientation in 3D space not only adds more parameters, increasing the complexity of the model, but also complicates the lifting from 2D rotations into 6D orientations.

\subsection{Semi-supervised Training}
Semi-supervised learning combines a small amount of labeled data with a large amount of unlabeled data during training. 
In the context of pose representation, Generative Adversarial Networks (GANs) are valuable for data augmentation, helping to distinguish realistic poses from unrealistic ones in datasets where only 2D annotations are available. 
For instance,~\cite{GAN_pose_AIGN} employs GANs to work with unpaired 2D/3D datasets and includes a 2D projection consistency term to ensure accuracy.
\cite{GAN_pose_AL_wild} introduces a novel multi-source discriminator that differentiates between predicted 3D poses and ground-truth data from real-world images. 
Additionally,~\cite{semi_3d_to_2d_projection} describes a weakly supervised method for 3D pose estimation using an adversarial setup with a new Random Projection layer.
\cite{semi_Ordinal_Depth_Supervision} recommends using the ordinal depths of human joints as a minimal supervision signal to make use of the variety found in 2D human pose datasets. 
Simultaneously,~\cite{3D_PE_semi} details back-projection, an effective semi-supervised training method that utilizes unlabeled video data to train GCNs for 3D pose estimation using TCN.

%%%%%%%%%%%%%%%%%%%%%%%%%%%%%%%%%%%%%%%%%%%%%%%%%%%%%%%%%%%%%%%%%%%%%%%%

\section{Methodology}
\label{sec:method}

\begin{figure*}[ht]
    \centering
    \includegraphics[width=\linewidth]{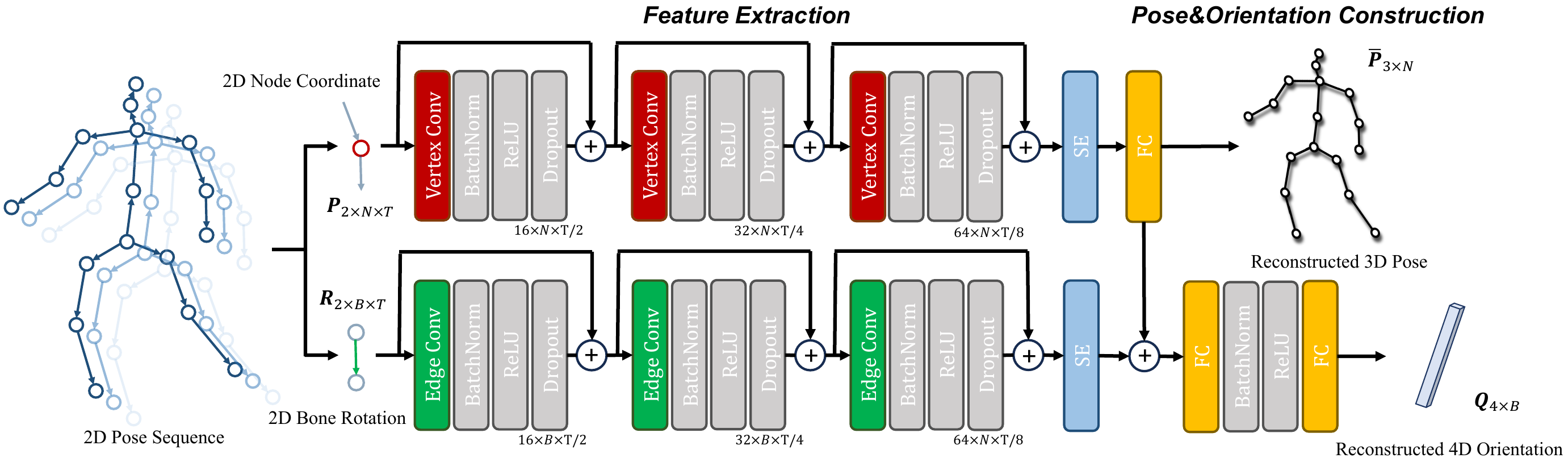}
    \vspace{3pt}
    \caption{Whole architecture of Q-GCN. Q-GCN begins by dividing the input 2D pose sequence into node coordinates and bone rotations, which are represented as vertices and edges in a directed graph, respectively. 
    It then extracts spatial-temporal features from these vertices and edges, incorporating a residual connection with each convolution operation. 
    Following this feature extraction, Q-GCN reconstructs the human 3D pose and 4D orientation using a fully-connected (FC) layer that includes a Squeeze and Excitation (SE) block. }
    \label{fig:Q-GCN}
\end{figure*}

This section introduces the whole structure of Quater-GCN (Q-GCN) and the semi-supervised training strategy we employ. 
Q-GCN extracts both temporal information from position of node joint and rotation from bone joint with a sequence, and reconstructs a more sophisticated human pose representation by 3D coordinates and orientations. 
Figure~\ref{fig:Q-GCN} shows the whole architecture of the model. 
The input of Q-GCN are the 2D keypoint coordinates estimated from a video, and the 2D bone joint rotations derived from these coordinates.
The output of Q-GCN are the 3D positions of each node joint and the orientations of each bone joint. 

\subsection{Graph Configuration}

\begin{figure}[ht]
    \centering
    \includegraphics[width=0.8\linewidth]{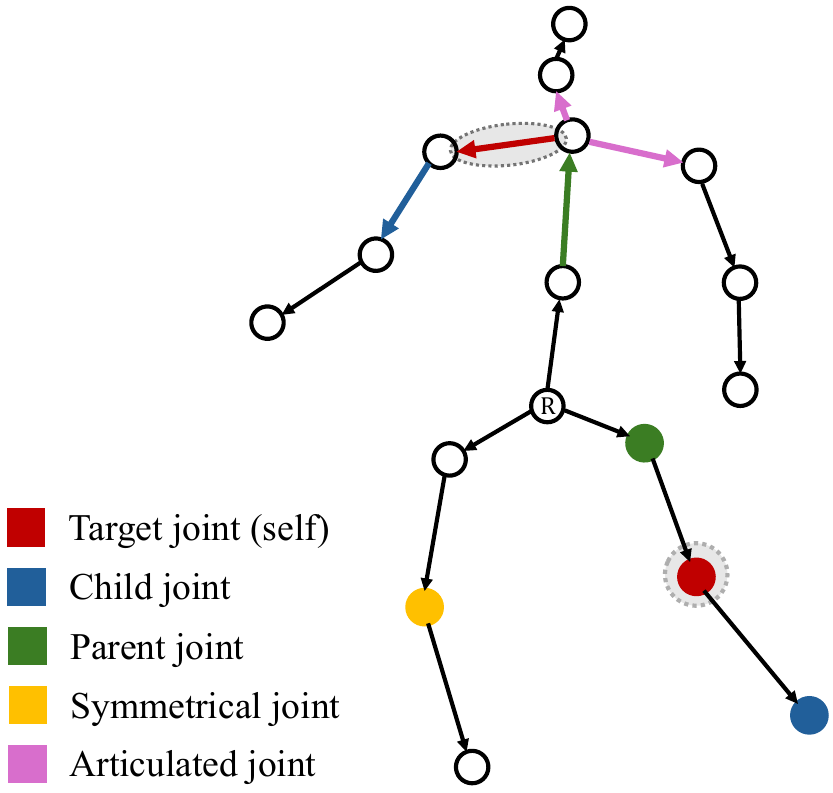}
    \vspace{3pt}
    \caption{Whole configuration of the directed graph and the sampling strategy in Q-GCN. In this graph, vertices and edges are organized in a hierarchical structure, with the root node (typically the pelvis node) serving as the initialization point. The sampling strategy is illustrated by marking the target vertex and edge joint with a dot circle, while different colors on the joints indicate the subsets they belong to, as defined by the sampling strategy.}
    \label{fig:config}
\end{figure}

In 2D space, node joint positions can be represented as coordinate sequence within a video as 
$P=\{ p_{t,n} \in \mathbb{R}^2 | t=1,2,\ldots,T; n=1,2,\ldots,N\}$, 
where $T$ and $N$ represent the number of frames in the sequence and the number of node joints in the human skeleton, respectively. 
We also incorporate a sequence of rotations for each bone joint within the 2D coordinate space, expressed as
$R=\{ r_{t,b} \in \mathbb{R}^2 | t=1,2,3,\ldots,T; b=1,2,3,\ldots,B\}$, 
where $B$ denotes the number of bone joints, and $r_{t,b}$ contains a 2-tuple of the cosine and sine values of the rotation angle $\theta \in [-\pi, \pi ]$ for each bone joint $b$, starting from the initial position.
All rotations are defined within the Local Coordinate System (LCS). 
In LCS, the origin point is established at the parent node of each bone, with the initial position ($0$ degrees) aligned with the direction in which the parent bone extends from this origin. 
Utilizing LCS is crucial as it preserves the spatial relationships between a bone joint and its parent.

Despite with the ignorance of the kinematic dependencies between joint and bones from the previous GCNs, we represent the skeleton data of human pose as a directed acyclic graph (DAG) inspired by~\cite{DGCN}, with each key node joint as vertex while each key bone joint as edge, as shown in Figure~\ref{fig:config}. 
We define this directed graph as
$G = \{(v_{t}, e_{t}) | t=1,2,\ldots,T \}$, 
where $v_{t} = \{ v_{t,n} | t=1,2,\ldots,T; n=1,2,\ldots,N\}$, and $e_{t} = \{ e_{t,b} | t=1,2,\ldots,T; b=1,2,3,\ldots,B\}$ represent the sets of key node joints and bone joints respectively.
The features of vertex $v_{t,n}$ and edge $e_{t,b}$ are initialized with their respective 2D coordinates $p_{t,n}$ and rotations $r_{t,b}$.

Q-GCN predicts both 3D positions and orientations to reconstruct human pose representation.
The 3D coordinates of the node joints in a human pose are denoted as
$\Bar{P}=\{ \Bar{p}_{n} \in \mathbb{R}^3 | n=1,2,\ldots,N\}$. 
In addition, we represent the output orientations in 3D space using 4D Quaternions, expressed as
$Q = \{q_{b} \in \mathbb{R}^4 | b=1,2,3,\ldots,B\}$ 
along with the root coordinate in 3D space, $\Bar{p}_{root} \in \mathbb{R}^3$. 
Quaternions are preferred because they not only avoid Gimbal lock but require fewer dimensions compared to a 6D matrix~\cite{6D}.
Similar to the rotations in the 2D system, orientations in 3D space are defined within the LCS. 
The relative orientation of a child bone joint to its parent is calculated using the Quaternions in the World Coordinate System (WCS) as follows:
\begin{equation}
    {q}_{child|LCS} = Inv({q}_{parent|WCS}) \times {q}_{child|WCS}
\end{equation}
where $Inv({q})$ denotes the inverse of the Quaternion ${q}$, and $\times$ denotes Quaternion multiplication. 
This design enhances the understanding of the internal dynamics and dependencies exerted from parent node to child node, as established by prior research~\cite{CDGCN}.

\subsection{Feature Extraction}
Similar to~\cite{model12_stgcn}, we first implement a basic spatial-temporal graph convolution block to extract the features both for positions of node and rotations of bones within the graph. 

\subsubsection{Sampling Strategy}
We define a neighbor set $\mathcal{B}_{n}^{v}$ as a spatial graph convolutional filter for vertex $v_{t,n}$, and set $\mathcal{B}_{b}^e$ for edge $e_{t,b}$.
Consequently, for the convolutional filter of vertex, we define four distinct neighbor subsets: 
(1) the vertex itself;
(2) the subset of parent vertices, which includes vertices that directly point to the target vertex (closer to root vertex);
(3) the subset of child vertices, which comprises vertices directly pointed by the target vertex; and 
(4) the subset of symmetrical vertices.
The inclusion of the symmetrical vertex subset addresses the issue of pendant vertices (also known as leaf vertices), such as the left or right hand, which do not have child vertex. 
Relying solely on the feature extraction of its parent vertex can result in a poor global representation~\cite{pe_glagcn}.
In addition, for the convolutional filter for edges, similar to vertices, we also define four distinct neighbor subsets:
(1) the edge itself;
(2) the subset of parent edge, which includes edges directly point to the target vertex;
(3) the subset of child edges, which are the edges directly pointed by the target edge; and
(4) the subset of articulating edges.
The inclusion of the articulating edge subset addresses the specific needs of edges like the left (or right) shoulder or neck (pink bone joints in Figure~\ref{fig:config}). 
These bone joints start from the root node and are articulated with one another, sharing close spatial dependencies.
Therefore, the kernel size $K$ is set to 4 both for vertex and edge filters, corresponding to the 4 subsets. 

To implement the subsets, mappings $h_{t,n}^{v} \rightarrow \{ 0,\dots, K-1\}$ and $h_{t,b}^{e}\rightarrow \{ 0,\dots, K-1\}$ are used to index each subset with a numeric label. 
Therefore, this convolutional operations of vertex and edge can be written as 
\begin{equation}
    f_{out}^{v}(v_{t,n}) = \sum_{v_{t,n'} \in \mathcal{B}_{n}^{v}} \frac{1}{Z_{t,n'}} f_{in}^{v}(v_{t,n'})W^{v}(h_{t,n}^{v}(v_{t,n'}))
    \label{equ:1}
\end{equation}

\begin{equation}
    f_{out}^{e}(e_{t,b}) = \sum_{e_{t,b'} \in \mathcal{B}_{b}^{e}} \frac{1}{Z_{t,{b'}}} f_{in}^{e}(e_{t,{b'}})W^{e}(h_{t,b}^{e}(v_{t,{b'}}))
    \label{equ:2}
\end{equation}
where the functions $f_{in}^{v}(v_{t,n'}): v_{t,n'} \rightarrow \mathbb{R}^2$ and $f_{in}^{e}(e_{t,{b'}}): e_{t,{b'}} \rightarrow \mathbb{R}^2$ denote the mappings that retrieve the attribute features of neighbor node joint $v_{t,n'}$ and neighbor bone joint $e_{t,{b'}}$ of $v_{t,n}$ and $e_{t,b}$ respectively.  
Note that the attribute features encapsulate both position of nodes and rotation of bone joints. 
$Z_{t,n'}$ and $Z_{t,{b'}}$ serve as normalization factors, equal to the cardinality of their respective subsets.
The weight functions $W^v(h_{t,n}(v_{t,n'}))$ and $W^e(h_{t,b}(v_{t,{b'}}))$ correspond to the mappings for $\mathcal{B}_{n}^{v}$ and $\mathcal{B}_{b}^e$ respectively, which are implemented by indexing a $(2,K)$ tensor. 

\subsubsection{Dependency Representation}
Within a pose frame, the graph convolution, as determined by the sampling strategy, is consistently implemented using adjacency matrices~\cite{pe_glagcn, ar_ddgcn, CDGCN}. 
Accordingly, for a directed graph containing $N$ vertices and $B$ edges, we define an $N \times N$ adjacent matrix $\mathbf{A}^{v}$ for the vertices and a $B \times B$ adjacent matrices $\mathbf{A}^{e}$ for edges.
The elements of these matrices represent the relationships between the corresponding vertices or edges, facilitating the propagation of information through the graph based on these defined connections.

However, an adjacency matrix that lacks hierarchical spatial information is not adequate for representing the directed edges within a directed graph. 
Inspired by~\cite{DGCN}, we employ incidence matrices for both vertices and edges to address this limitation.
Furthermore, we define two $N \times B$ incidence matrices $\mathbf{P}^{v}$ and $\mathbf{C}^{v}$, where the elements indicate whether a given edge is the parent or child edge of a vertex.  
Similarly, we define two $B \times N$ incidence matrices $\mathbf{P}^{e}$ and $\mathbf{C}^{e}$, to specify whether a vertex is the parent or child of an edge. 
For instance, for a parent edge (or vertex) of a vertex $v_{n}$ (or an edge $e_{b}$), the corresponding element in the parent incidence matrix $\mathbf{P}^{v}$ (or $\mathbf{P}^{e}$) is set to $1$.
Conversely, for its child edge (or vertex), the corresponding element in the child incidence matrix $\mathbf{C}^{v}$ (or $\mathbf{C}^{e}$) is set to $1$, with all other elements set to $0$. 
This structure enhances the graph representation by clearly defining and utilizing the hierarchical relationships between vertices and edges within the data.

\subsubsection{Adaptive Representation}
Inspired by~\cite{li2018adaptive}, we also incorporate an adaptive design to enhance the flexibility of the ST-GCN block. 
Specifically, utilizing $K$ spatial sampling strategies, we employ the sum of the incidence matrices for vertices, $\sum_{k=0}^{K-1} A_{k}^{v}$, and for edges, $\sum_{k=0}^{K-1} A_{k}^{e}$.
This allows for the implementation of Equations~\ref{equ:1} and~\ref{equ:2} using these matrices as follows:
\begin{equation}
    \mathbf{H}_{t}^{v} = \sum_{k=0}^{K-1} [ \Bar{\mathbf{A}}_{k}^{v}, \Bar{\mathbf{P}}_{k}^{e}, \Bar{\mathbf{C}}_{k}^{e} ] \mathbf{F}_{k}^{v} \mathbf{W}_{k}^{v}
\end{equation}

\begin{equation}
    \mathbf{H}_{t}^{e} = \sum_{k=0}^{K-1} [ \Bar{\mathbf{A}}_{k}^{e}, \Bar{\mathbf{P}}_{k}^{v}, \Bar{\mathbf{C}}_{k}^{v} ] \mathbf{F}_{k}^{e} \mathbf{W}_{k}^{e}
\end{equation}
where $\Bar{\mathbf{A}}_{k} = \mathbf{\Lambda}_{k}^{\frac{1}{2}} \mathbf{A}_{k} \mathbf{\Lambda}_{k}^{\frac{1}{2}}$ represents the normalized adjacency matrix of $\mathbf{A}_{k}$ for both vertices and edges.
Following the approach used in~\cite{gcn}, $\mathbf{\Lambda}_{k}^{ii} = \sum_{n}(\Bar{\mathbf{A}}_{k}^{in}) + \alpha$ forms a diagonal matrix, with $\alpha$ set to $0.001$ to prevent empty rows. 
$[\cdot]$ denotes the concatenation operation. 
$\mathbf{W}_{k}$ represents the weighting function for Equations~\ref{equ:1} and~\ref{equ:2}, corresponding to a weight tensor of the $1\times1$ convolution operation. 
$\mathbf{F}_{k}$ specifies the attribute features of all the neighbor joints sampled into the subset $k$.
This structured approach facilitates comprehensive spatial-temporal feature extraction, essential for dynamic pose estimation tasks.

Therefore, each convolution layer in Q-GCN is implemented using a $1 \times T$ classical 2D convolution layer, where $T$ represents the temporal kernel size. 
The output from this layer, $\mathbf{H}_{t}$, is sequentially processed through a batch normalization layer, which is followed by a ReLU activation layer, and then a dropout layer, collectively forming a single convolutional block.
Additionally, a residual connection~\cite{resnet} is integrated in each convolution layer to enhance the learning process.

\subsection{Pose and Orientation Construction}

For pose and orientation construction, drawing inspiration from~\cite{se, PoseGraphNet}, we initially employ a Squeeze and Excitation (SE) block to recalibrate the channel-wise features for both coordinates and rotations. 
This enhances the model's sensitivity to informative features for both pose and orientation. 
Subsequently, we utilize a fully-connected layer to integrate multi-scale feature maps, which helps in predicting the final 3D poses with enhanced accuracy.
In terms of orientation, the process begins by concatenating the rotation features with the predicted 3D coordinates.
This concatenated output is then processed through two fully-connected layers.
Between these layers, we insert a batch normalization layer followed by a ReLU activation layer.
The complete architecture of Q-GCN is illustrated in Figure~\ref{fig:Q-GCN}.

\subsection{Semi-supervision Strategy}

\begin{figure}[ht]
    \centering
    \includegraphics[width=\linewidth]{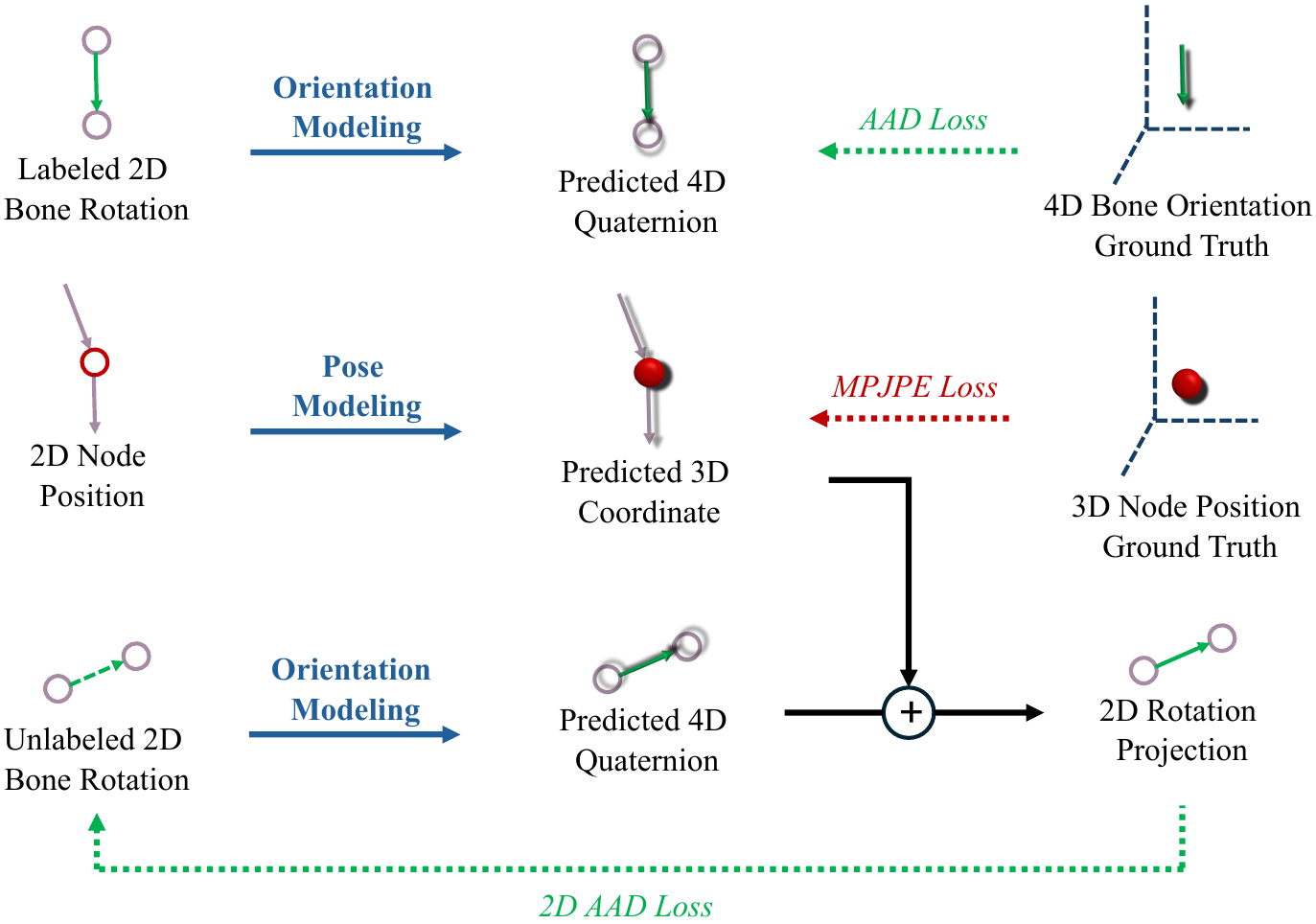}
    \vspace{3pt}
    \caption{Semi-supervised training strategy for orientation regression. The unlabeled 2D rotations in the latter half of the batch are regressed using projected 2D rotations. These projections are derived from combining the predicted 4D orientations with the predicted 3D positions, both of which are initially trained using labeled data during the first half of the batch. }
    \label{fig:semi}
\end{figure}

Inspired by~\cite{3D_PE_semi}, we introduce a semi-supervised training strategy to address the shortage of labeled ground truth of Quaternion regression of orientation.
Figure~\ref{fig:semi} illustrates this strategy, which integrates both supervised and unsupervised components.
Initially, in the supervised phase (first half batch), we feed Q-GCN with labeled 2D node coordinates and labeled 2D bone rotations.
For the loss functions, we utilize the mean per-joint position error (MPJPE) loss~\cite{3D_PE_semi} to regress the predicted 3D node coordinates based on the ground truth. 
To accurately regress the predicted 4D bone Quaternions, we develop an Average Angular Distance (AAD) loss function aimed at minimizing the angular distance between the ground truth and the predicted values:
\begin{equation}
    \mathcal{L}_{angular} = \frac{1}{T} \frac{1}{B} \sum_{t=1}^{T} \sum_{b=1}^{B} 2\arccos{(Re({\Bar{q}}_{t,b} \times conj({q}_{t,b}))}
\label{eq:AAD}
\end{equation}
where $\Bar{q}_{t,b}$ and $q_{t,b}$ represent the ground truth and predicted Quaternion of bone joint $b$ at frame $t$, respectively. 
The functions $Re(\cdot)$ and $conj(\cdot)$ return the real part and the conjugate of a Quaternion, respectively.

In the unsupervised phrase (last half batch), which deals with unlabeled 2D bone rotations and corresponding labeled coordinates, we first use the model, trained with the initial batch, to predict the 4D bone orientations and corresponding 3D node coordinates. 
Following this, we integrate the predicted 3D coordinates with the 4D Quaternions to facilitate the computation of the 2D rotation projection.
This setup allows us to regress the unlabeled data with these projections using a 2D AAD loss:
\begin{equation}
    \mathcal{L}_{angular}^{2D} = \frac{1}{T} \frac{1}{B} \sum_{t=1}^{T} \sum_{b=1}^{B} |\Bar{\theta}_{t,b}-\theta_{t,b}|
\end{equation}
Where ${\Bar{\theta}}_{t,b}$ and ${\theta}_{t,b}$ denote the ground truth and the predicted rotation angle of bone joint $b$ at frame $t$.

%%%%%%%%%%%%%%%%%%%%%%%%%%%%%%%%%%%%%%%%%%%%%%%%%%%%%%%%%%%%%%%%%%%%%%%%

\section{Experimental Results}
\label{sec:results}
\subsection{Datasets and Evaluation}
Our method is assessed using three public datasets, with \textsl{Human3.6M}~\cite{h36m} and \textsl{HumanEva-I}~\cite{humaneva} focusing on human major-part keypoints, while \textsl{H3WB}~\cite{h3wb} on human whole-body keypoints.
Consistent with established practices in prior research~\cite{pe_glagcn, 3D_PE_semi, CDGCN}, our training on \textsl{Human3.6M} involves data from subjects S1, S5, S6, S7, and S8,  with testing conducted on subjects S9 and S11. 
For \textsl{HumanEva-I}, we use data depicting the actions ``walk'' and ``jog'' performed by subjects S1, S2, and S3, applying it to both training and testing. 
In the case of \textsl{H3WB}, we adhere to the settings outlined in~\cite{h3wb}, utilizing a training set of 80k 2D-3D pairs and testing on half of the total available test samples.

Our evaluation protocols include the Mean Per-Joint Position Error (MPJPE) and the Pose-aligned MPJPE (P-MPJPE), also referred to as $Protocol\#1$ and $Protocol\#2$, respectively. 
For \textsl{Human3.6M}, both protocols are implemented, whereas for \textsl{HumanEva}, we solely apply $Protocol\#2$. 
In addition, considering the sparse research on orientation evaluation~\cite{quaternet}, we empoly our mean Average Angular Distance (mAAD) loss for assessment, as detailed in Equation~\ref{eq:AAD}.

\subsection{Implementation Details}
For 2D pose detection, we apply the methodologies utilized in \textsl{Human3.6M} and \textsl{HumanEva} as detailed in~\cite{CDGCN}, using CPN~\cite{CPN} and MRCNN~\cite{RN004} respectively for detection. 
Additionally, we conduct experiments with ground truth (GT) 2D pose detection for all three datasets, noting that H36W is evaluated solely on the GT 2D whole-body pose due to its extensive keypoint coverage.

Regarding model settings, we adapt the sizes of the graph convolutional filters to match the structure of the 2D pose, with filters set to accommodate 17 node and 16 bone joints in \textsl{Human3.6M} and 16 node and 15 bone joints in \textsl{HumanEva}, respectively. 
For \textsl{H3WB}, due to the extensive number of keypoints, we categorize the whole-body keypoints into distinct groups: body, face, left hand, and right hand. 
Each group's filter size in our model is specifically tailored to match the number of node and bone joints associated with that particular body part.
For example, in the configuration for the right hand within H3WB, the filters for vertex and edge are set to 21 and 20, respectively, corresponding to the 21 node joints and 20 bone joints that comprise the right hand.
To assess the efficacy of the proposed model, particularly in orientation construction and semi-supervised learning, additional ablation experiments are performed on the \textsl{Human3.6M} dataset.

In terms of hyperparameters, batch sizes are set at 512 for \textsl{Human3.6M}, 256 for \textsl{HumanEva-I}, and 128 for \textsl{H3WB}, in line with~\cite{pe_glagcn, h3wb}. 
Consistent with~\cite{pe_glagcn}, the ranger optimizer is used, and the model is trained using the MPJPE loss for 80 epochs for \textsl{Human3.6M} and 1000 epochs for \textsl{HumanEva-I}, starting with an initial learning rate of 0.01. 
The dropout rate is maintained at 0.25, and data augmentation via horizontal flipping is applied during both training and testing phases. 

\subsection{Comparable Results}
\begin{table*}[ht]
  \centering
  \resizebox{\linewidth}{!}{%
  \begin{tabular}{lcccccccccccccccc}
    \toprule[2pt]
	Method  &  Dir.  &  Disc.  &  Eat.  &  Greet  &  Phone  &  Photo  &  Pose  &  Purch.  &  Sit  &  SitD.  &  Smoke  &  Wait  &  WalkD.  &  Walk  &  WalkT.  &  Avg.  \\
    \midrule
Martinez et al. \cite{RN011} (ICCV’17)   &  51.8  &  56.2  &  58.1  &  59.0  &  69.5  &  78.4  &  55.2  &  58.1  &  74.0  &  94.6  &  62.3  &  59.1  &  65.1  &  49.5  &  52.4  &  62.9  \\
Fang et al. \cite{RN034} (AAAI’18)  &  50.1  &  54.3  &  57.0  &  57.1  &  66.6  &  73.3  &  53.4  &  55.7  &  72.8  &  88.6  &  60.3  &  57.7  &  62.7  &  47.5  &  50.6  &  60.4  \\
Pavlakos et al. \cite{semi_Ordinal_Depth_Supervision} (CVPR’18)   &  48.5  &  54.4  &  54.4  &  52.0  &  59.4  &  65.3  & 49.9  &  52.9  &  65.8  &  71.1  &  56.6  &  52.9  &  60.9  &  44.7  &  47.8  &  56.2  \\
Lee et al. \cite{RN036} (ECCV’18) †    &  40.2  &  49.2  &  47.8  &  52.6  &  50.1  &  75.0  &  50.2  &  43.0  &  55.8  &  73.9  &  54.1  &  55.6  &  58.2  &  43.3  &  43.3  &  52.8  \\
Zhao et al. \cite{pr_semgcn} (CVPR’19) *   &  47.3  &  60.7  &  51.4  &  60.5  &  61.1  &  49.9  &  47.3  &  68.1  &  86.2  &  \textbf{55.0}  &  67.8  &  61.0  &  \underline{42.1}  &  60.6  &  45.3  &  57.6  \\
Ci et al. \cite{RN029} (ICCV’19)   &  46.8  &  52.3  &  44.7  &  50.4  &  52.9  &  68.9  &  49.6  &  46.4  &  60.2  &  78.9  &  51.2  &  50.0  &  54.8  &  40.4  &  43.3  &  52.7  \\
Pavllo et al. \cite{3D_PE_semi} (CVPR’19) †   &  45.2  &  46.7  &  43.3  &  45.6  &  48.1  &  55.1  &  44.6  &  44.3  &  57.3  &  65.8  &  47.1  &  44.0  &  49.0  &  32.8  &  33.9  &  46.8  \\
Cai et al. \cite{RN038} (ICCV’19) † *  &  44.6  &  47.4  &  45.6  &  48.8  &  50.8  &  59.0  &  47.2  &  43.9  &  57.9  &  61.9  &  49.7  &  46.6  &  51.3  &  37.1  &  39.4  &  48.8  \\
Xu et al. \cite{RN015} (CVPR’20) † &   \textbf{37.4}  &   43.5  &  42.7  &  42.7  &  46.6  &  59.7  &  41.3  &  45.1  &  \underline{52.7}  &  60.2  &  45.8  &  43.1  &  47.7  &  33.7  &  37.1  &  45.6  \\
Liu et al. \cite{RN013} (CVPR’20) † &  41.8  &  44.8  &  41.1  &  44.9  &  47.4  &  54.1  &  43.4  &  42.2  &  56.2  &  63.6  &  45.3  &  43.5  &  45.3  &  31.3  &  32.2  &  45.1  \\
Zeng et al. \cite{zeng2020srnet} (ECCV’20) †  &  46.6  &  47.1  &  43.9  &  41.6  &  45.8  &  49.6  &  46.5  &  40.0  &  53.4  &  61.1  &  46.1  &  42.6  &  43.1  &  31.5  &  32.6  &  44.8  \\
Xu and Takano \cite{RN039} (CVPR’21) *  &  45.2  &  49.9  &  47.5  &  50.9  &  54.9  &  66.1  &  48.5  &  46.3  &  59.7  &  71.5  &  51.4  &  48.6  &  53.9  &  39.9  &  44.1  &  51.9  \\
Zhou et al. \cite{RN040} (PAMI’21) †  & \underline{38.5}  &  45.8  &  \underline{40.3}  &  54.9  &  \textbf{39.5}  &  \textbf{45.9}  &  \textbf{39.2}  &  43.1  &  \textbf{49.2}  &  71.1  &  \textbf{41.0}  &  53.6  &  44.5  &  33.2  &  34.1  &  45.1  \\
Li et al. \cite{trans_Mhformer} (CVPR’22) †    &  39.2  &  \underline{43.1}  &  \textbf{40.1}  &  \textbf{40.9}  &  \underline{44.9}  &  51.2  &  \underline{40.6}  &  41.3  &  53.5  &  60.3  &  43.7  &  41.1  &  43.8  &  29.8  &  30.6  &  \underline{43.0}  \\
Shan et al. \cite{shan2022p} (ECCV’22) †  &  38.9  &  \textbf{42.7}  &  40.4  &  \underline{41.1}  &  45.6  &  \underline{49.7}  &  40.9  &  \textbf{39.9}  &  55.5  &  \underline{59.4}  &  44.9  &  42.2  &  \textbf{42.7}  &  \underline{29.4}  &  \underline{29.4}  &  \textbf{42.8}  \\
Yu et al. \cite{pe_glagcn} (ICCV'23) † * &  41.3  &  44.3  &  40.8  &  41.8  &  45.9  &  54.1  &  42.1  &  41.5  &  57.8  &  62.9  &  45.0  &  \underline{42.8}  &  45.9  &  \underline{29.4}  &  29.9  &  44.4  \\
\rowcolor{grayrow}
Our Q-GCN (T=243, CPN) † * &  41.1  &  43.3  &  40.4  &  41.3  &  \underline{44.9}  &  53.2  &  41.7  &  \underline{41.1}  &  54.9  &  65.2  &  \underline{43.5}  &  \textbf{41.3}  & \textbf{42.7}  &  \textbf{29.1}  &  \textbf{29.2}  &  43.5  \\

\midrule
Martinez et al. \cite{RN011} (ICCV’17)  &  37.7  &  44.4  &  40.3  &  42.1  &  48.2  &  54.9  &  44.4  &  42.1  &  54.6  &  58.0  &  45.1  &  46.4  &  47.6  &  36.4  &  40.4  &  45.5  \\
Lee et al. \cite{RN036} (ECCV’18) †  &  32.1  &  36.6  &  34.3  &  37.8  &  44.5  &  49.9  &  40.9  &  36.2  &  44.1  &  45.6  &  35.3  &  35.9  &  30.3  &  37.6  &  35.5  &  38.4  \\
Zhao et al. \cite{pr_semgcn} (CVPR’19)   &  37.8  &  49.4  &  37.6  &  40.9  &  45.1  &  41.4  &  40.1  &  48.3  &  50.1  &  42.2  &  53.5  &  44.3  &  40.5  &  47.3  &  39.0  &  43.8  \\
Ci et al. \cite{RN029} (ICCV’19)  &  36.3  &  38.8  &  29.7  &  37.8  &  34.6  &  42.5  &  39.8  &  32.5  &  \underline{36.2}  &  {39.5}  &  34.4  &  38.4  &  38.2  &  31.3  &  34.2  &  36.3  \\
Liu et al. \cite{RN013} (CVPR’20) † &  34.5  &  37.1  &  33.6  &  34.2  &  32.9  &  37.1  &  39.6  &  35.8  &  40.7  &  41.4  &  33.0  &  33.8  &  33.0  &  26.6  &  26.9  &  34.7  \\
Xu and Takano \cite{RN039} (CVPR’21) *  &  35.8  &  38.1  &  31.0  &  35.3  &  35.8  &  43.2  &  37.3  &  31.7  &  38.4  &  45.5  &  35.4  &  36.7  &  36.8  &  27.9  &  30.7  &  35.8  \\
Zheng et al. \cite{zheng20213d} (ICCV’21) †   &  30.0  &  33.6  &  29.9  &  31.0  &  30.2  &  33.3  &  34.8  &  31.4  &  37.8  &  38.6  &  31.7  &  31.5  &  29.0  &  23.3  &  23.1  &  31.3  \\
Li et al. \cite{trans_Mhformer}  (CVPR’22) †  &  {27.7}  &  32.1  &  {29.1}  &  28.9  &  30.0  &  33.9  &  33.0  &  31.2  &  37.0  &  39.3  &  30.0  &  31.0  &  29.4  &  22.2  &  23.0  &  30.5  \\
Shan et al. \cite{shan2022p} (ECCV’22) † &  28.5  &  {30.1}  &  \textbf{28.6}  &  {27.9}  &  {29.8}  &  {33.2}  &  {31.3}  &  {27.8}  &  \textbf{36.0}  &  \textbf{37.4}  &  \underline{29.7}  &  {29.5}  &  \underline{28.1}  &  \underline{21.0}  &  {21.0}  &  \underline{29.3}  \\
Yu et al. (ICCV'23) † * &  \underline{26.5}  &  \underline{27.2}  &  29.2  &  \textbf{25.4}  &  \textbf{28.2}  &  \underline{31.7}  &  \textbf{29.5}  &  \underline{26.9}  &  37.8  &  39.9  &  {29.9}  &  \textbf{27.0}  &  \textbf{27.3}  &  \textbf{20.5}  &  \textbf{20.8}  &  \textbf{28.5}  \\
\rowcolor{grayrow}
Our Q-GCN (T=243, GT) † * &  \textbf{26.1}  &  \textbf{26.8}  &  \underline{28.8}  &  \underline{26.1}  &  \underline{28.5}  &  \textbf{31.1}  &  \underline{29.9}  &  \textbf{26.4}  &  37.1  &  \underline{39.4}  &  \textbf{29.6}  &  \underline{28.1}  &  {28.3}  &  \underline{20.7}  &  \textbf{20.2}  &  \textbf{28.5}  \\
    \bottomrule[2pt]
  \end{tabular}}
  \vspace{3pt}
   \caption{Reconstruction error on \textsl{Human3.6M} under $Protocol\#1$. Top table: 2D pose sequences detected by CPN as input. Bottom table: 2D pose sequences with GT as input. (†) uses temporal information. (*) uses GCN-based model. Lower is better, best in bold, second best underlined. }
  \label{table:h36m_1}  
\end{table*}

\begin{table*}[ht]
  \centering
  \resizebox{\linewidth}{!}{%
  \begin{tabular}{lcccccccccccccccc}
    \toprule[2pt]
	Method  &  Dir.  &  Disc.  &  Eat.  &  Greet  &  Phone  &  Photo  &  Pose  &  Purch.  &  Sit  &  SitD.  &  Smoke  &  Wait  &  WalkD.  &  Walk  &  WalkT.  &  Avg.  \\
    \midrule
Martinez et al. \cite{RN011} (ICCV’17)  &  39.5  &  43.2  &  46.4  &  47.0  &  51.0  &  56.0  &  41.4  &  40.6  &  56.5  &  69.4  &  49.2  &  45.0  &  49.5  &  38.0  &  43.1  &  47.7  \\
Fang et al. \cite{RN034} (AAAI’18)  &  38.2  &  41.7  &  43.7  &  44.9  &  48.5  &  55.3  &  40.2  &  38.2  &  54.5  &  64.4  &  47.2  &  44.3  &  47.3  &  36.7  &  41.7  &  45.7  \\
Pavlakos et al. \cite{semi_Ordinal_Depth_Supervision} (CVPR’18)  &  34.7  &  39.8  &  41.8  &  38.6  &  42.5  &  47.5  &  38.0  &  36.6  &  50.7  &  56.8  &  42.6  &  39.6  &  43.9  &  32.1  &  36.5  &  41.8  \\
Lee et al. \cite{RN036} (ECCV’18) †  &  34.9  &  {35.2}  &  43.2  &  42.6  &  46.2  &  55.0  &  37.6  &  38.8  &  50.9  &  67.3  &  48.9  &  35.2  &  \textbf{31.0}  &  50.7  &  34.6  &  43.4  \\
Cai et al. \cite{RN038} (ICCV’19) † * &  35.7  &  37.8  &  36.9  &  40.7  &  39.6  &  45.2  &  37.4  &  34.5  &  46.9  &  50.1  &  40.5  &  36.1  &  41.0  &  29.6  &  33.2  &  39.0  \\
Pavllo et al. \cite{3D_PE_semi} (CVPR’19) † &  34.1  &  36.1  &  34.4  &  37.2  &  36.4  &  42.2  &  34.4  &  33.6  &  45.0  &  52.5  &  37.4  &  33.8  &  37.8  &  25.6  &  27.3  &  36.5  \\
Xu et al. \cite{RN015} (CVPR’20) † &  \textbf{31.0}  &  \textbf{34.8}  &  34.7  &  {34.4}  &  36.2  &  43.9  &  \underline{31.6}  &  33.5  &  \textbf{42.3}  &  {49.0}  &  37.1  &  33.0  &  39.1  &  26.9  &  31.9  &  36.2  \\
Chen et al. \cite{CPN} (ICCV’20) †  &  32.9  &  {35.2}  &  35.6  &  {34.4}  &  36.4  &  42.7  &  \textbf{31.2}  &  {32.5}  &  45.6  &  50.2  &  37.3  & 32.8  &  36.3  &  26.0  &  23.9  &  35.5  \\
Liu et al. \cite{RN013} (CVPR’20) † &  {32.3}  &  {35.2}  &  {33.3}  &  35.8 &  35.9  &  \textbf{41.5}  &  33.2  &  32.7  &  44.6  &  50.9  &  37.0  &  \underline{32.4}  &  37.0  &  25.2  &  27.2  &  35.6  \\
Shan et al. \cite{shan2021improving} (MM’21) †  &  32.5  &  36.2  &  33.2  &  35.3  &  \underline{35.6}  &  \underline{42.1}  &  32.6  &  \underline{31.9}  &  \underline{42.6}  &  \textbf{47.9}  &  {36.6}  &  \textbf{32.1}  &  {34.8}  &  {24.2}  &  {25.8}  &  {35.0}  \\
Shan et al. \cite{shan2022p} (ECCV’22) † &  31.3  &  35.2  &  32.9  &  33.9  &  35.4  &  39.3  &  32.5  &  31.5  &  44.6  &  \underline{48.2}  &  \underline{36.3}  &  32.9  &  \underline{34.4}  &  \underline{23.8}  &  \textbf{23.9}  &  \textbf{34.4}  \\
Yu et al. \cite{pe_glagcn} (ICCV'23) † * &  32.4  &  35.3  &  \underline{32.6}  &  \underline{34.2}  &  \textbf{35.0}  &  \underline{42.1}  &  32.1  &  \underline{31.9}  &  45.5  &  49.5  &  \textbf{36.1}  &  \underline{32.4}  &  35.6  &  \textbf{23.5}  &  {24.7}  &  {34.8}  \\
    \rowcolor{grayrow}
Our Q-GCN (T=243, CPN) † * &  \underline{31.1}  &  \underline{34.9}  &  \textbf{32.4}  &  \textbf{33.7}  &  {36.3}  &  {42.8}  &  \underline{31.6}  &  \textbf{31.2}  &  44.7  &  {48.6}  &  {36.9}  &  \underline{32.4}  &  35.4  &  {24.1}  &  \underline{24.4}  &  \underline{34.7}  \\

    \midrule
  Martinez et al.  \cite{RN011} (ICCV’17)  &  -  &  -  &  -  &  -  &  -  &  -  &  -  &  -  &  -  &  -  &  -  &  -  &  -  &  -  &  -  &  37.1  \\
Ci et al. \cite{RN029} (ICCV’19)   &  {24.6}  &  {28.6}  &  {24.0}  &  {27.9}  &  {27.1}  &  {31.0}  &  {28.0}  &  {25.0}  &  {31.2}  &  {35.1}  &  {27.6} &  {28.0}  &  {29.1}  &  {24.3}  &  {26.9}  &  {27.9}  \\
Yu et al.~\cite{pe_glagcn} (ICCV'23) † * &  \underline{20.2}  &  \underline{21.9}  &  \underline{21.7}  &  \textbf{19.9}  &  \underline{21.6}  &  \underline{24.7}  &  \underline{22.5}  &  \underline{20.8}  &  \underline{28.6}  &  \textbf{33.1}  &  \underline{22.7}  &  \textbf{20.6}  &  \underline{20.3}  &  \underline{15.9}  &  \underline{16.2}  &  \underline{22.0}  \\
    \rowcolor{grayrow}
Our Q-GCN (T=243, GT) † * &  \textbf{20.1}  &  \textbf{21.4}  &  \textbf{21.5}  &  \underline{20.3}  &  \textbf{21.1}  &  \textbf{24.2}  &  \textbf{21.2}  &  \textbf{20.3}  &  \textbf{27.2}  &  \underline{34.2}  &  \textbf{21.5}  &  \underline{21.4}  &  \textbf{20.0}  &  \underline{14.8}  &  \textbf{15.1}  &  \textbf{21.6}  \\

        \bottomrule[2pt]
  \end{tabular}}
  \vspace{3pt}
   \caption{Reconstruction error after rigid alignment on \textsl{Human3.6M} under $Protocol\#2$. Top table: 2D pose sequences detected by CPN as input. Bottom table: 2D pose sequences with GT as input. (†) uses temporal information. (*) uses GCN-based model. Lower is better, best in bold, second best underlined.}
  \label{table:h36m_2}  
\end{table*}

Tables~\ref{table:h36m_1} and~\ref{table:h36m_2} showcase comparisons of Q-GCN against state-of-the-art (SOTA) methods on the \textsl{Human3.6M} dataset under $Protocol\#1$ and $Protocol\#2$, respectively. 
For the \textsl{HumanEva} dataset, Table~\ref{table:eva_2} displays results under $Protocol\#2$ alongside other SOTA methods. 
Overall, Q-GCN surpasses these methods in terms of average results across both evaluation protocols, achieving the lowest average MPJPE loss on \textsl{Human3.6M} 
with 2D GT pose as input, and the lowest P-MPJPE loss on \textsl{Human3.6M} and on \textsl{HumanEva} when using both detected 2D poses (by MRCNN) and 2D GT poses.
Particularly with GT pose inputs, Q-GCN demonstrates substantial performance improvements across nearly all action classes. 
Additionally, GLA-GCN~\cite{pe_glagcn}, which also incorporates temporal information within a GCN-based model, achieves superior outcomes comparing to other methods. 
This highlights the crucial contribution of temporal information and GCN-based architecture in capturing the dynamics of human movement over time and the relational dependencies between body parts, essential for accurate 3D pose estimation.
However, as noted in the top sections of Tables~\ref{table:h36m_1} and~\ref{table:h36m_2}, actions such as Discussion (Disc), Taking Photos (Photo), Posing (Pose), and Sitting (Sit) tend to exhibit higher errors with methods leveraging temporal information and GCN-based architecture, especially when using detected 2D poses. 
This may be due to the lower movement amplitude of these actions compared to others like Walking Dog (WalkD.), Walking (Walk), and Walking Together (WalkT.), which are easier for GCN-based models to capture due to their more pronounced dynamic features.

To gain deeper insights into how our method performs on different body parts, we conducted 2D-to-3D lifting experiments on human whole-body keypoints. 
Table~\ref{tab:h3wb} presents results comparing our Q-GCN method to state-of-the-art (SOTA) methods on the \textsl{H3WB} dataset, evaluated using $Protocol\#1$. 
Overall, Q-GCN consistently achieves top performance, ranking within the top two for the lowest error across all tests, indicating its strong capability in modeling complex pose structures with numerous vertices and edges.
Additionally, Table~\ref{tabel:maad} details the performance of various methods in constructing 4D orientation, evaluated using mean Average Angular Distance (mAAD). 
Q-GCN outperforms all competing methods, demonstrating superior effectiveness in accurately modeling every body part.

\begin{table}[ht]
  \centering
    \resizebox{\linewidth}{!}{%
  \begin{tabular}{lccccccc}
    \toprule[2pt]
   \multirow{2}[3]{*}{Method} & \multicolumn{3}{c}{Walk} & \multicolumn{3}{c}{Jog} &\multirow{2}[3]{*}{Avg} \\
    \cmidrule(lr){2-4} \cmidrule(lr){5-7}
    & S1 & S2 & S3  & S1 & S2 & S3  \\
    \midrule
Martinez et al. \cite{RN011} (ICCV’17)  &  19.7  &  17.4  &  46.8  &  26.9  &  18.2  &  18.6  &  24.6  \\ 
Fang et al. \cite{RN034} (AAAI’18)  &  19.4  &  16.8  &  37.4  &  30.4  &  17.6  &  16.3  &  23.0  \\ 
Pavlakos et al. \cite{semi_Ordinal_Depth_Supervision} (CVPR’18)   &  18.8  &  12.7  &  29.2  &  23.5  &  15.4  &  14.5  &  19.0  \\ 
Lee et al. \cite{RN036} (ECCV’18) †  &  18.6  &  19.9  &  30.5  &  25.7  &  16.8  &  17.7  &  21.5  \\ 
Pavllo et al. \cite{3D_PE_semi} (CVPR’19) †   &  13.9  &  10.2  &  46.6  &  20.9  &  13.1  &  13.8  &  19.8  \\ 
Liu et al. \cite{RN013} (CVPR’20) † & 13.1  &  {9.8}  &  {26.8}  &  \textbf{16.9}  &  \underline{12.8}  &  {13.3}  &  {15.5}  \\ 
Zheng et al. \cite{zheng20213d} (ICCV’21) † &  14.4  &  10.2  &  46.6  &  22.7  &  13.4  &  13.4  &  20.1  \\ 
Li et al. \cite{trans_temporal_contexts} (TMM’22) † &  14.0  &  10.0  &  32.8  &  19.5  &  13.6  &  14.2  &  17.4  \\ 
Zhang et al. \cite{trans_Seq2seq} (CVPR’22) †  &   {12.7}  &  10.9  &  \textbf{17.6}  &  22.6  &  15.8  &  17.0  &  16.1 \\ 
Yu et al. \cite{pe_glagcn} (ICCV'23) † * &  \underline{12.5}  &  \textbf{9.1}  &  26.9  &  {18.5}  &  \textbf{12.7}  &  \underline{12.8}  &  \underline{15.4}  \\ 
\rowcolor{grayrow}
Our Q-GCN (T=27, MRCNN) † * &  \textbf{12.1}  &  \underline{9.3}  &  \underline{25.4}  &  \underline{17.9}  &  {12.9}  &  \textbf{12.7}  &  \textbf{15.1}  \\ 
\midrule
Li et al. \cite{trans_temporal_contexts} (TMM’22) † &  {9.7}  &  {7.6}  & {15.8}  &  {12.3}  &  {9.4}  &  {11.2}  &  {11.1}  \\ 
Yu et al. \cite{pe_glagcn} (ICCV'23) † * &   \underline{8.7}  &   \underline{6.8}  &   \underline{11.5}  &   \textbf{10.1}  &   \textbf{8.2}  &   \underline{9.9}  &   \underline{9.2}  \\ 
\rowcolor{grayrow}
Our Q-GCN (T=27, GT) † *  &   \textbf{8.5}  &   \textbf{6.4}  &   \textbf{10.8}  &   \underline{10.7}  &   \underline{8.9}  &   \textbf{9.6}  &   \textbf{9.1}  \\ 
    \bottomrule[2pt]
  \end{tabular}}
  \vspace{2pt}
  \caption{Reconstruction error after rigid alignment on \textsl{HumanEva} under $Protocol\#2$. Top table: 2D pose sequences detected by MRCNN as input. Bottom table: 2D pose sequences with GT as input. (†) uses temporal information. (*) uses GCN-based model. Lower is better, best in bold, second best underlined.}
	\label{table:eva_2}
\end{table}

\begin{table}[ht]
    \centering
    \resizebox{\linewidth}{!}{
    \begin{tabular}{l c c c c}
     \toprule[2pt]
    \multicolumn{1}{l}{Method} & \multicolumn{1}{c}{All} &
    \multicolumn{1}{c}{Body} &
    \multicolumn{1}{c}{Face / aligned$^\dagger$} &
    \multicolumn{1}{c}{Hand / aligned$^\ddagger$} \\
    \midrule
    % \textit{\textsl{H3WB}} & & &  \\
    SMPL-X~\cite{SMPL-X} & 188.9 & 166.0 & 208.3 / 23.7 & 170.2 / 44.4 \\
    CanonPose\cite{cannopose}$^*$ & 186.7 & 193.7 & 188.4 / 24.6 & 180.2 / 48.9 \\
    SimpleBaseline \cite{RN011}$^*$ & 125.4 & 125.7 & 115.9 / 24.6 & 140.7 / 42.5 \\
    CanonPose\cite{cannopose} \textit{w} 3D sv.$^*$ & 117.7 & 117.5 & 112.0 / 17.9 & 126.9 / 38.3 \\
    Large SimpleBaseline\cite{RN011}$^*$ & 112.3 & 112.6 & 110.6 / \textbf{14.6} & \textbf{114.8} / \textbf{31.7} \\
    Jointformer~\cite{jointformer} & \underline{88.3} & \underline{84.9} & \underline{66.5} / 17.8 & 125.3 / 43.7 \\
    \rowcolor{grayrow}
    Our Q-GCN & \textbf{82.4} & \textbf{79.6} & \textbf{63.2} / \underline{15.4} & \underline{119.6} / \underline{39.2} \\
     \bottomrule[2pt]
    \end{tabular}
    }
    \vspace{2pt}
    \caption{Reconstruction error w/o rigid alignment on \textsl{H3WB} under $Protocol\#1$. (*) output normalized predictions. (Sv.) for supervision. Lower is better, best in bold, second best underlined. MPJPE metric in mm. All results are pelvis aligned, except $\dagger$ and $\ddagger$ show nose and wrist aligned results for face and hands, respectively.}
    \label{tab:h3wb}
\end{table}

\begin{table}[ht]
\begin{center}
\resizebox{\linewidth}{!}{
\begin{tabular}{l c c c c c}
\toprule[2pt]
{Method} & {All} & {Major-part} & {Upper-body} & {Lower-body} &{Hands}\\
% \toprule[2pt]
\toprule[1pt]
SMPL-X~\cite{SMPL-X} &     123 & 72 & 89 & 64 & 167\\
Jointformer~\cite{jointformer} & \underline{77} & 66 & 72 & 49 & 103\\
GLA-GCN~\cite{pe_glagcn}  &     {79} & \underline{54} & \underline{63} & \underline{41} & \underline{91}\\
\rowcolor{grayrow}
Our Q-GCN  & \textbf{67} & \textbf{32} & \textbf{41} & \textbf{27} & \textbf{83}\\
\bottomrule[2pt]
\end{tabular}
}
\end{center}
\vspace{2pt}
\caption{Reconstruction error on \textsl{H3WB} under mAAD loss, scaled by $10^{3}$. Lower is better, best in bold, second best underlined.}
\label{tabel:maad}
\end{table}

\subsection{Ablation Studies}

The ablation studies were conducted from two perspectives: the effect of different receptive fields and a component-wise comparison.

Table~\ref{table:field} compares various SOTA methods on the \textsl{Human3.6M} dataset, applying different receptive fields to GT 2D poses under $Protocol\#1$. 
Generally, the results indicate that a larger receptive field tends to yield better performance across all methods. 
Notably, Q-GCN outperforms other methods, particularly when utilizing larger receptive fields. 
However, with a smaller receptive field (e.g., $T=27$), GCN-based models show lower performance compared to transformer-based models like~\cite{trans_Mhformer}, likely due to the expansive receptive field afforded by the attention mechanism of transformer.

Table~\ref{tab:keydesign} details an ablation study on key component designs of Q-GCN, focusing on orientation construction, semi-supervised training strategies, and the use of directed graphs (implemented via incidence matrices). 
We established a baseline model that includes only vertex convolution and pose construction using an undirected graph, regressed by 3D coordinates. 
The addition of orientation entails incorporating edge convolution and orientation construction regressed by 4D Quaternions. 
"With semi-supervision" indicates the application of the semi-supervised training strategy, which can only be implemented alongside orientation construction. 
"With directed graph" refers to the implementation of directed graph convolution network.
The results demonstrate that the fully-equipped Q-GCN significantly outperforms the other configurations, confirming the effectiveness of the designed components. 
Comparing setups $\#2$ and $\#3$, it is evident that the semi-supervised training strategy effectively addresses the lack of Quaternion annotations for orientation regression. 
Additionally, comparison between setups $\#3$ and $\#5$ shows that with both including orientation construction, the impact of semi-supervision is more significant than that of using a directed graph. 
Comparisons among $\#1$, $\#2$, and $\#4$ suggest that implementing orientation construction yields more substantial benefits than using a directed graph. 
These findings further underscore the importance of orientation information in accurate 3D human pose estimation.
\begin{table}[ht]
  \centering
    % \resizebox{0.98\linewidth}{!}
    {%
  \begin{tabular}{lcc}
    \toprule[2pt]
Medhod  &  Frames   &  MPJPE (mm)  \\
\midrule 
Pavllo et al. \cite{3D_PE_semi} (CVPR’19) † &  $T$ = 27   &  40.6  \\
Liu et al. \cite{RN013} (CVPR’20) † &   $T$ = 27  &   38.9  \\
Li et al. \cite{trans_Mhformer} (CVPR’22) † &   $T$ = 27 &  \textbf{34.3}  \\
Yu et al.~\cite{pe_glagcn} (ICCV'23) † * &   $T$ = 27  &  \underline{34.4}  \\
  \rowcolor{grayrow}
Our Q-GCN † * &   $T$ = 27  &  {34.8}  \\ % valid ~50.3
\midrule 
Pavllo et al. \cite{3D_PE_semi} (CVPR’19) †  &   $T$ = 81  &   38.7  \\
Liu et al. \cite{RN013} (CVPR’20) † &   $T$ = 81   &  36.2  \\
Li et al. \cite{trans_Mhformer} (CVPR’22) † &   $T$ = 81  &   {32.7}  \\
Yu et al.~\cite{pe_glagcn} (ICCV'23) †  *&   $T$ = 81  &  \textbf{31.5}  \\ % valid ~48.7
  \rowcolor{grayrow}
Our Q-GCN † *&   $T$ = 81  &  \underline{31.9}  \\ % valid ~48.7
\midrule 
Pavllo et al. \cite{3D_PE_semi} (CVPR’19) † &   $T$ = 243 &  37.8  \\
Liu et al. \cite{RN013} (CVPR’20) † &   $T$ = 243   &  {34.7}  \\
Zhang et al. \cite{trans_Seq2seq} (CVPR’22) †   &   $T$ = 243   &  \underline{21.6}  \\ 
Yu et al.~\cite{pe_glagcn} (ICCV'23) † *&   $T$ = 243  &   {28.5} \\
  \rowcolor{grayrow}
Our Q-GCN † * &   $T$ = 243   &  \textbf{21.3}  \\  
\bottomrule[2pt]
  \end{tabular}}
  \vspace{2pt}
  \caption{Comparison with state-of-the-art methods on \textsl{Human3.6M} under $Protocol\#1$, implemented with different receptive fields of ground truth 2D pose. (*) uses GCN model.}
	\label{table:field}
\end{table}

\begin{table}[H]
  \centering
    \resizebox{0.98\linewidth}{!}{%
  \begin{tabular}{llcccc}
    \toprule[2pt]
  \multirow{2}[3]{*}{\#} & \multirow{2}[3]{*}{Method} & \multicolumn{2}{c}{\textsl{Human3.6M}} & \multicolumn{2}{c}{\textsl{HumanEva-I}} \\
    \cmidrule(lr){3-4} \cmidrule(lr){5-6}
  &  & CPN & GT   & MRCNN & GT  \\
    \midrule
1 & Baseline  &  47.1  &  36.5  &  23.4  &  11.6  \\ 
2 & With orientation ($\mathcal{L}_{angular}$)  &  40.2  &  30.1  &  20.9  & 10.7    \\ 
3 & With orientation \& semi-supervision   &  \underline{36.0}  &  \underline{27.2}  &  \underline{17.9}  &  10.1   \\
4 & With directed graph &  45.2  &  {31.7}  &  21.5  &  11.2   \\
5 & With orientation \& directed graph &  39.6  &  {29.7}  &  19.6  &  \underline{9.8}   \\
%\midrule
\rowcolor{grayrow}
6 & Q-GCN (With all)   &  \textbf{34.7}  &  \textbf{21.6}  &  \textbf{15.1}  &  \textbf{9.1}  \\ 
    \bottomrule[2pt]
  \end{tabular}}
  \vspace{3pt}
  \caption{Ablation study on key designs of our Q-GCN. The results are based on the average value of $Protocol\#2$ implemented with $27$ receptive fields for various 2D pose detections of the \textsl{Human3.6M} and \textsl{HumanEva-I}.}
	\label{tab:keydesign}
\end{table}
% \vspace{-5pt}
%%%%%%%%%%%%%%%%%%%%%%%%%%%%%%%%%%%%%%%%%%%%%%%%%%%%%%%%%%%%%%%%%%%%%%%%
\section{Conclusion}
\label{sec:conclusion}
In this paper, we tackle the shortcomings of current 3D pose estimation methods that focus on spatial position but overlook the orientation or rotation of bones, which are crucial for understanding poses in 3D space. 
We introduce Quater-GCN, a novel graph convolutional network for 2D-to-3D pose lifting that incorporates both orientation and position data. 
This approach is further refined by a semi-supervised training strategy for 4D Quaternion regression, providing a more sophisticated pose representation. 
Rigorous evaluations across public datasets show that Q-GCN consistently outperforms state-of-the-art methods, especially with ground truth 2D poses, demonstrating its robustness and accuracy. 
Ablation studies highlight key components like orientation regression, directed graphs, and semi-supervised learning as significant contributors to our system's performance. 
This paper emphasizes the importance of integrating orientation data and semi-supervised learning to enhance 3D human pose estimation.

%%%%%%%%%%%%%%%%%%%%%%%%%%%%%%%%%%%%%%%%%%%%%%%%%%%%%%%%%%%%%%%%%%%%%%%%
\newpage

\bibliography{mybibfile}

\clearpage

\setcounter{page}{1}
\setcounter{section}{0}

\newpage
\twocolumn[
\centering
\Large
\vspace{0.5em}\textbf{**Appendix**}\\
\vspace{1.0em}
] 

\renewcommand\thesection{\Alph{section}}

\section{Extended Introduction to Q-GCN}

This section offers supplementary details to further elucidate the Q-GCN introduction.

\subsection{Dataset for Orientation Regression}
We employ coordinate transformation to convert the 3D coordinates of discrete joint nodes within the human body skeleton into the rotations of each bone joint in. 
Additionally, we articulate these bone joints using 3D vectors.
For example, the vector representing the upper arm is derived by subtracting the coordinates of the shoulder node from those of the elbow node: 
\begin{equation}
\Vec{v}=\boldsymbol{p}_{elbow}-\boldsymbol{p}_{shoulder}
\end{equation}

\begin{figure}[ht]
\centering
\includegraphics[width=0.8\columnwidth]{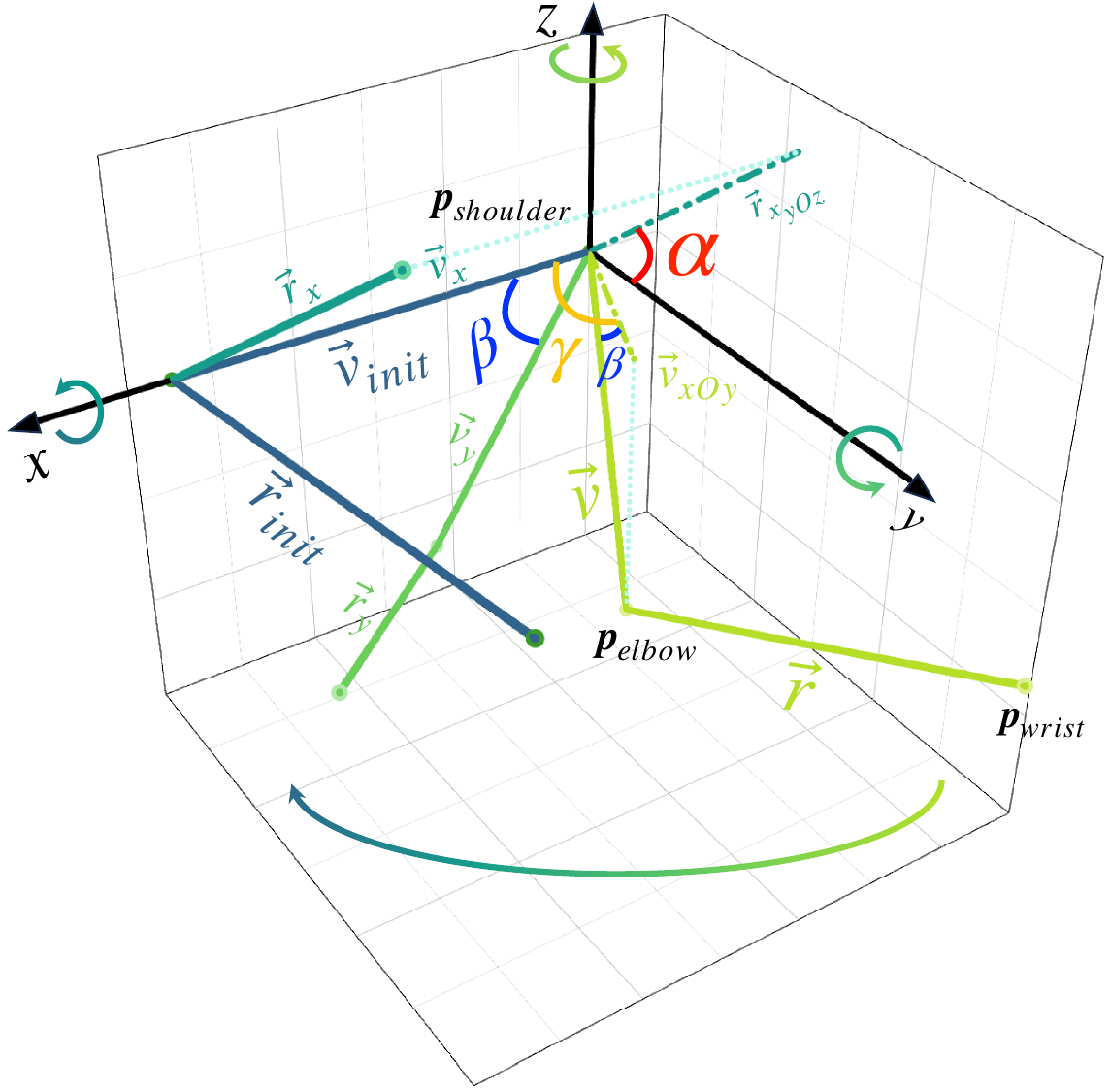} 
\caption{Schematic diagram of Euler angle calculation:
$\boldsymbol{p}_{shoulder}$ (rotation center), $\boldsymbol{p}_{elbow}$, and $\boldsymbol{p}_{wrist}$ represent the nodes corresponding to the shoulder, elbow and wrist joints in 3D World Coordinate System respectively. 
Given the initial target vector $\Vec{v}_{init}$ and initial reference vector $\Vec{r}_{init}$, and proceeding to rotate them sequentially around the $x,y \text{ and } z$ axes by angles $\alpha,\beta, \text{ and } \gamma$ respectively, $(\Vec{v}_x,\Vec{r}_x)$, $(\Vec{v}_y,\Vec{r}_y)$ and finally $(\Vec{v},\Vec{r})$ can be obtained. 
The diagram displays a gradient arrow, signifying the Euler angle calculation order, which is in reverse to the rotation sequence.
Consequently, we calculate $\gamma$ as the rotation angle from the positive $x$-axis to $\vec{v}_{xOy}$, which representing the projection of vector $\vec{v}$ onto the $xOy$ plane. 
Likewise, $\beta$ is inferred as the rotation angle between vector $\vec{v}$ and the $xOy$ plane. 
Lastly, $\alpha$ is calculated as the angle spanning from the positive $y$-axis to $\vec{r}_{x_{yOz}}$, the projection of $\vec{r}_{x}$ onto the $yOz$ plane.}
\label{euler_angle}
\end{figure}

Therefore, poses can be articulated by the rotations of each bone vector from the initial position to the current position. 
Initially, We compute the static Euler angles of each vector in World Coordinate System in $x-y-z$ sequence, utilizing the corresponding coordinates of skeleton nodes. 
To illustrate, Figure~\ref{euler_angle} demonstrates the calculation of the Euler angles $(\alpha,\beta,\gamma)$ for the upper arm vector $\Vec{v}$, with shoulder joint node $\boldsymbol{p}_{shoulder}$ as the center of rotation. 
Hence, Euler angles $\gamma$ and $\beta$ can be calculated as:
\begin{equation}
    \vec{v}_{xOy}=\vec{v}-\langle \vec{v},\vec{n}_{z} \rangle 
\end{equation}
\begin{equation}
    \gamma=\arccos{
    \frac{\vec{v}_{xOy} \cdot \vec{i}_{x}}
    {||\vec{v}_{xOy}||}    
}
\end{equation}
\begin{equation}
    \beta=\arccos{
    \frac{\vec{v}_{xOy} \cdot \vec{v}}
    {||\vec{v}_{xOy}|| \cdot ||\vec{v}||}   
    }
\end{equation}
where $\vec{n}_{z}$ represents the normal vector of $xOy$ plane and $\vec{i}_{x}$ signifies the unit vector along $x$-axis. 

Nonetheless, when a 3D vector is parallel to the $x$-axis, it becomes incapable of representing a rotation about the $x$-axis.
To address this, we introduce a reference vector $\vec{r}$, represented by the vector formed by the child bone joint of the target bone joint, such as the forearm (child) and upper arm (parent), aiding in the calculation of $\alpha$. 
Consequently, $\alpha$ can be calculated using $\vec{r}$, $\gamma$, and $\beta$ as follows:

\begin{equation}
    \vec{r}_{x}=RM^{-1}_{z}(\gamma) \cdot RM^{-1}_{y}(\beta) \cdot \vec{r}
\end{equation}
\begin{equation}
    \alpha=\arccos{
    \frac{(\vec{r}_{x}-\langle \vec{r}_{x},\vec{n}_{x} \rangle ) \cdot \vec{i}_{y}}
    {||\vec{r}_{x}-\langle \vec{r}_{x},\vec{n}_{x} \rangle ||}    
}
\end{equation}
where $\vec{r}_{x}$ represents the vector $\vec{r}$ rotated around $x$-axis by angle $\alpha$. $RM^{-1}_{z}(\gamma)$ and $RM^{-1}_{y}(\beta)$ denote the inverse rotation metrics for angles $\gamma$ and $\beta$ respectively. $\vec{n}_{x}$ denotes the normal vector to $yOz$ plane, and $\vec{i}_{y}$ is the unit vector along $y$-axis.

However, to avoid the potential Gimbal Lock issue associated with Euler angles, we convert these angles into Quaternions\footnotemark[1] for rotation representation. 
The transformation from Euler angles to Quaternions is illustrated below:
\begin{equation}
    \boldsymbol{q}=q_{x}\boldsymbol{i}+q_{y}\boldsymbol{j}+q_{z}\boldsymbol{k}+q_{w}
\end{equation}
\begin{equation}
    q_{x}=\sin{(\theta/2)}\cdot \cos{\alpha} 
\end{equation}
\begin{equation}
    q_{y}=\sin{(\theta/2)}\cdot \cos{\beta} 
\end{equation}
\begin{equation}
    q_{z}=\sin{(\theta/2)}\cdot \cos{\gamma} 
\end{equation}
\begin{equation}
    q_{w}=\cos{(\theta/2)} 
\end{equation}
where $\theta$ represents the rotation angle transitioning from the initial position to the current position\footnotemark[1], also serves as the composition of rotations by angles $\alpha$, $\beta$ and $\gamma$ in 3D coordinate system.

\section{Experiment Configurations}

This section provides comprehensive details about the experimental setups mentioned in the main paper.

\subsection{Experimental Environments}
This section introduces the experimental environments used for the evaluations.
\begin{itemize}
    \item \textbf{Wisteria/BDEC-01 Supercomputer System}:\\
    CPU: Intel$^\circledR$ Xeon$^\circledR$ Gold 6258R CPU @2.70GHz;\\
    CPU number: 112;\\
    Memory: 512 GiB;\\
    GPU: NVIDIA A100-SXM4-40GB;\\
    GPU number: 8;\\
    System: Red Hat Enterprise Linux 8;\\
    Python: 3.8.18; \\
    NVCC: V11.6.124;\\
    GCC: 8.3.1 20191121 (Red Hat 8.3.1-5);\\
    PyTorch: 1.13.1;\\
    CUDA: 11.6;\\
    CuDNN: 8.3.2.\\ 

    \item \textbf{Server}:\\
    CPU: Intel$^\circledR$ Xeon$^\circledR$ W-2125 CPU @4.00GHz;\\
    CPU number: 8;\\
    Memory: 128 GB;\\
    GPU: Quadro RTX 6000;\\
    GPU number: 1;\\
    System: Ubuntu 20.04.6 LTS;\\
    Python: 3.7.16; \\
    NVCC: V11.3.58;\\
    GCC: (Ubuntu 9.4.0-1ubuntu1~20.04.2) 9.4.0;\\
    PyTorch: 1.11.0+cu113;\\
    CUDA: 11.3;\\
    CuDNN: 8.2.\\ 

    \item \textbf{PC}:\\
    CPU: Intel63 Family 6 Model 158 Stepping 9 GenuineIntel @3.6GHz;\\
    CPU number: 1;\\
    Memory: 32 GB;\\
    GPU: NVIDIA GeForce RTX 3060;\\
    GPU number: 1;\\
    System: Microsoft Windows 10.0.19045;\\
    Python: 3.8.17; \\
    NVCC: V11.8.89;\\
    GCC: N/A;\\
    PyTorch: 2.0.0;\\
    CUDA: 11.8;\\
    CuDNN: 8.7.\\ 

\end{itemize}

\section{Project Code}
We attach out project code with this pdf file. 
For further evaluation on Q-GCN, please refer to the \textbf{``code''} folder.
For detailed information, please refer to the \textbf{``README.md''} attached with code.

\section{Data}
Due to the license agreements concerning the use of the Human3.6M, NTU RGB+D, and H3WB datasets, we are unable to distribute any datasets derived from these sources. 

\end{document}